\documentclass[11pt,letterpaper,usenames,dvipsnames]{article}
\usepackage{eacl2017}
\usepackage[utf8x]{inputenc}
\usepackage{times}
\usepackage{url}
\usepackage{latexsym}
\usepackage{expex}
\usepackage{booktabs}
\usepackage{amsmath}
\usepackage{amssymb}
\usepackage{adjustbox}
\usepackage{tikz}
\usepackage{ifthen}
\usepackage{ellipsis}

\usetikzlibrary{positioning}
\usetikzlibrary{fit}
\usetikzlibrary{shadows}
\usepackage{algorithm2e}
\usepackage{subcaption}
\usepackage{multirow}
\usepackage{ragged2e}
\usepackage{pgfplots}
\PrerenderUnicode{ä}
\def\checkmark{\tikz\fill[scale=0.4](0,.35) -- (.25,0) -- (1,.7) -- (.25,.15) -- cycle;}

\title{Imagination improves Multimodal Translation}

\author{ 
  Desmond Elliott$^{\star\diamond}$ \and Ákos Kádár$^{\dagger}$ \\
  $^{\star}$ILLC, University of Amsterdam\\
  $^{\diamond}$School of Informatics, University of Edinburgh \\
  $^{\dagger}$TiCC, Tilburg University \\
  {\tt d.elliott@ed.ac.uk}, {\tt a.kadar@uvt.nl}}

\date{}
\eaclfinalcopy

\begin{document}
\maketitle

\begin{abstract}

We decompose multimodal translation into two sub-tasks: learning to translate and learning visually grounded representations. In a multitask learning framework, translations are learned in an attention-based encoder-decoder, and grounded representations are learned through image representation prediction. Our approach improves translation performance compared to the state of the art on the Multi30K dataset. Furthermore, it is equally effective if we train the image prediction task on the external MS COCO dataset, and we find improvements if we train the translation model on the external News Commentary parallel text.

\end{abstract}
\section{Introduction}

Multimodal machine translation is the task of translating sentences in context, such as images paired with a parallel text \cite{Specia2016}. This is an emerging task in the area of multilingual multimodal natural language processing. Progress on this task may prove useful for translating the captions of the images illustrating online news articles, and for multilingual closed captioning in international television and cinema.

Initial efforts have not convincingly demonstrated that visual context can improve translation quality. 
In the results of the First Multimodal Translation Shared Task, only three systems outperformed an off-the-shelf text-only phrase-based machine translation model, and the best performing system was equally effective with or without the visual features \cite{Specia2016}. There remains an open question about how translation models should take advantage of visual context.

\definecolor{myred}{RGB}{228,26,28}
\definecolor{myorange}{RGB}{255,127,0}
\definecolor{mypurple}{RGB}{152,78,163}
\definecolor{mygreen}{RGB}{77,175,74}
\definecolor{myblue}{RGB}{55,126,184}
\definecolor{mymagenta}{RGB}{231,41,138}
\tikzstyle{every pin edge}=[<-,shorten <=1pt]
\tikzstyle{neuron}=[circle,fill=black!25,minimum size=8pt,inner sep=0pt]
\tikzstyle{visual neuron}=[neuron, draw=myred, fill=myred]
\tikzstyle{source neuron}=[neuron, draw=mypurple, fill=mypurple]
\tikzstyle{sentence neuron}=[neuron, draw=myblue, fill=myblue]
\tikzstyle{hidden neuron}=[neuron, draw=mygreen, fill=mygreen]
\tikzstyle{output neuron}=[neuron, draw=gray, fill=gray]
\tikzstyle{word neuron}=[neuron, draw=myred, fill=myred]
\tikzstyle{dense neuron}=[neuron, draw=mypurple, fill=mypurple]
\tikzstyle{empty word neuron}=[neuron, draw=myred, fill=white]
\tikzstyle{annot} = [text width=6em, text centered]
\tikzstyle{connection}=[opacity=1]
\tikzstyle{conn}=[opacity=1, draw=gray, line width=0.1ex, shorten >= 0.15 cm, shorten <= 0.15cm]
\tikzstyle{sconn}=[opacity=1, draw=gray, line width=0.1ex, shorten >= 0.15 cm, shorten <= 0.05cm]
\tikzstyle{fithidneuron}=[draw=mygreen, rounded corners=8pt]
\tikzstyle{sourcefit}=[draw=myblue, rounded corners=8pt]
\tikzstyle{visualfit}=[draw=myred, rounded corners=8pt]

\begin{center}
\begin{figure}
\centering
\adjustbox{max width=0.45\textwidth}{
\begin{tikzpicture}[font=\Large\bfseries]

\def\initX{1}
\def\initY{1}
\def\layerD{3}
\def\times{5}
\def\imgt{10}
\def\dtimes{3}
\def\dinitX{5}
\def\dinitY{4}
\def\src{{"A","girl","eats","a","pancake"}}%
\def\tgt{{"","Ein","Mädchen"}}%

    \foreach \time / \x in {1,...,\times}
      {
      \foreach \name / \y in {1,...,\layerD} {
         \node[source neuron] (H\time-\name) at (\initX+2.25+\x, 0.5*\y*\initY) {};
        }
        \node[fithidneuron] (HiddenFit\time) [fit= (H\time-1) (H\time-3)] {};
    \node[below=2em of HiddenFit\time] (x\time) {\normalsize \strut \pgfmathparse{\src[\time-1]}\pgfmathresult};
    \draw[conn, ->] (x\time) to (HiddenFit\time);   
    }

    \node [below=4em of HiddenFit3] (enclabel) {\Large Shared Encoder};
    \node[sourcefit, line width=0.4ex] (SourceFit) [fit=(HiddenFit1) (HiddenFit5)] {};

    \draw [conn, ->] (HiddenFit1.40) to (HiddenFit2.140);
    \draw [conn, ->] (HiddenFit4.40) to (HiddenFit5.140);
    \draw [conn, ->] (HiddenFit2.40) to (HiddenFit3.140);
    \draw [conn, ->] (HiddenFit3.40) to (HiddenFit4.140);
    \draw [conn, <-] (HiddenFit1.320) to (HiddenFit2.220);
    \draw [conn, <-] (HiddenFit4.320) to (HiddenFit5.220);
    \draw [conn, <-] (HiddenFit2.320) to (HiddenFit3.220);
    \draw [conn, <-] (HiddenFit3.320) to (HiddenFit4.220);

    \node [source neuron, above=3em of HiddenFit3] (s-attn) {};
    \node [annot, above=0.5ex of s-attn, text width=12em] {\large Attention};
    \draw [sconn, ->] ([yshift=1ex]HiddenFit1.90) to (s-attn);
    \draw [sconn, ->] ([yshift=1ex]HiddenFit2.90) to (s-attn);
    \draw [sconn, ->] ([yshift=1ex]HiddenFit3.90) to (s-attn);
    \draw [sconn, ->] ([yshift=1ex]HiddenFit4.90) to (s-attn);
    \draw [sconn, ->] ([yshift=1ex]HiddenFit5.90) to (s-attn);

    \foreach \name / \y in {1,...,\layerD} {
       \node[sentence neuron] (M-\name) at (\initX+2.25+7, 0.5*\y*\initY) {};
    }
    \node[sourcefit] (SFit) [fit= (M-1) (M-3)] {};
    \node [annot, above=0.8ex of SFit, text width=4em] {\large \strut Average Pool};
    \node [below right=4.5ex and -4ex of SFit, text width=14ex] (enclabel) {\Large \begin{center}\centering  {\sc imaginet} Decoder\end{center}};

    \foreach \name / \y in {1,...,\layerD} {
       \node[visual neuron] (V-\name) at (\initX+2.25+9, 0.5*\y*\initY) {};
    }
    \node[visualfit] (VFit) [fit= (V-1) (V-3)] {};
    \draw [conn, ->] (SFit) to (VFit);
    \draw [conn, ->, draw=myblue] (SourceFit) to (SFit);
    \node [annot, above=0.5ex of VFit, text width=6em] {\large \strut Image};



      \foreach \time / \x in {1,...,\dtimes}
      {
        \foreach \name / \y in {1,...,\layerD}
        {  
           \node[hidden neuron] (DH\time-\name) at (\dinitX+\x*2, \dinitY+0.5*\y) {};
        }
        \node[fithidneuron] (DFit\time) [fit= (DH\time-1) (DH\time-3)] {};
        \ifthenelse{\x > 1}{\node[above=2em of DFit\time, text height=2ex] (y\x) {\small \strut \pgfmathparse{\tgt[\time-1]}\pgfmathresult};}{}
        \ifthenelse{\x > 1}{\draw[conn, ->] (DFit\time) to (y\time);}{}
      }

      \draw [conn, ->] (DFit1) to (DFit2);
      \draw [conn, ->] (DFit2) to (DFit3);
      \draw [conn, ->] (s-attn) to (DFit3.270);
      \draw [conn, ->] (y2) to (DFit3);

   \node [above=0.5ex of y2] (declabel) {\Large Translation Decoder};
\end{tikzpicture}
}
\caption{The Imagination model learns visually-grounded representations by sharing the encoder network between the Translation Decoder with image prediction in the {\sc imaginet} Decoder.}\label{fig:model}
\end{figure}
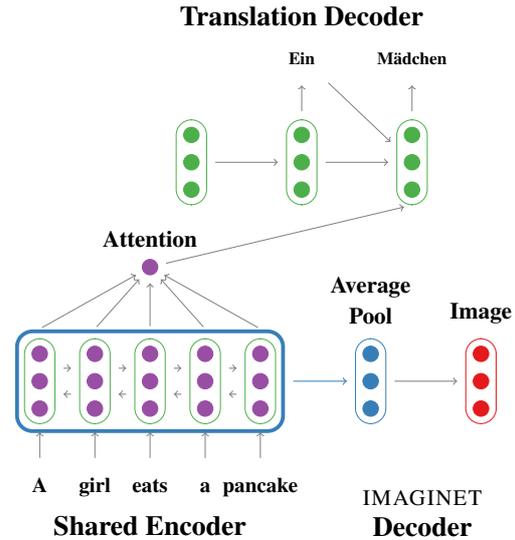
\end{center}

We present a multitask learning model that decomposes multimodal translation into learning a translation model and learning visually grounded representations. This decomposition means that our model can be trained over external datasets of parallel text or described images, making it possible to take advantage of existing resources. Figure \ref{fig:model} presents an overview of our model, Imagination, in which source language representations are shared between tasks
through the Shared Encoder. The translation decoder is an
attention-based neural machine translation model \cite{Bahdanau2015}, and the
image prediction decoder is trained to predict a global feature vector of an
image that is associated with a sentence \cite[{\sc imaginet}]{Chrupala2015}. This decomposition encourages grounded learning in the shared encoder because the {\sc imaginet} decoder is trained to imagine the image associated with a sentence.  
It has been shown that grounded representations are qualitatively different from their text-only counterparts \cite{Kadar2016} and correlate better with human similarity judgements \cite{Chrupala2015}.
We assess the success of the grounded learning 
 by evaluating the image prediction model on an image--sentence ranking task to determine if the shared representations are useful for image retrieval \cite{Hodosh2013}. In contrast with most previous work, our model does not take images as input at translation time, rather it learns grounded representations in the shared encoder.

We evaluate Imagination on the Multi30K dataset \cite{ElliottFrankSimaanSpecia2016} using a combination of in-domain and out-of-domain data. In the in-domain experiments, we find that multitasking translation with image prediction is competitive with the state of the art. Our model achieves 55.8 Meteor as a single model trained on multimodal in-domain data, and 57.6 Meteor as an ensemble.

In the experiments with out-of-domain resources, we find that the improvement in translation quality holds when training the {\sc imaginet} decoder on the MS COCO dataset of described images \cite{Chen2015}.
Furthermore, if we significantly improve our text-only baseline using out-of-domain parallel text from the News Commentary corpus \cite{Tiedemann2012}, we still find improvements in translation quality from the auxiliary image prediction task. Finally, we report a state-of-the-art result of 59.3 Meteor on the Multi30K corpus when ensembling models trained on in- and out-of-domain resources.\footnote{The implementation will be available upon publication.}

The main contributions of this paper are:
\begin{itemize}
\item We show how to apply multitask learning to multimodal translation. This makes it possible to train models for this task using external resources alongside the expensive triple-aligned source-target-image data.
 
\item We decompose multimodal translation into two tasks: learning to translate and learning grounded representations. We show that each task can be trained on large-scale external resources, e.g. parallel news text or images described in a single language.

\item We present a model that achieves state of the art results without using images as an input. Instead, our model learns visually grounded source language representations using an auxiliary image prediction objective. Our model does not need any additional parameters to translate unseen sentences.

 
\end{itemize}
\section{Problem Formulation}\label{sec:problem}

Multimodal translation is the task of producing target language translation $y$, given the source language sentence $x$ and additional context, such as an image $v$ \cite{Specia2016}. Let $x$ be a source language sentence consisting $N$ tokens: x$_1$, x$_2$, $\ldots$, x$_n$ and let $y$ be a target language sentence consisting $M$ tokens: y$_1$, y$_2$, $\ldots$, y$_m$. The training data consists of tuples $\mathcal{D} \in (x, y, v)$, where $x$ is a description of image $v$, and $y$ is a translation of $x$. 

Multimodal translation has previously been framed as minimising the negative log-likelihood of a translation model that is additionally conditioned on the image, i.e. $J(\theta) = - \sum_{j} \text{log p}(y_j|y_{<j}, x, v)$. Here, we decompose the problem into learning to translate and learning visually grounded representations. The decomposition is based on sharing parameters $\theta$ between these two tasks, and learning task-specific parameters $\phi$. We learn the parameters in a multitask model with shared parameters in the source language encoder. The translation model has task-specific parameters $\phi^t$ in the attention-based decoder, which are optimized through the translation loss $J_{T}(\theta, \phi^t)$. 
Grounded representations are learned through an image prediction model with task-specific parameters $\phi^g$ in the image-prediction
decoder by minimizing $J_{G}(\theta, \phi^g)$. The joint objective is given by mixing the translation and image prediction tasks with the parameter $w$:
\begin{align}
	J(\theta, \phi) = w J_{T}(\theta, \phi^{t}) +  (1 - w) J_G(\theta, \phi^{g})
\end{align}
Our decomposition of the problem makes it straightforward to optimise this objective without paired tuples, e.g. where we have an external dataset of described images $\mathcal{D}_{image} \in (x, v)$ or an external parallel corpus $\mathcal{D}_{text} \in (x, y)$. 


We train our multitask model following the approach of \newcite{Luong2016}. We define a primary task and an auxiliary task, and a set of parameters $\theta$ to be shared between the tasks. A minibatch of updates is performed for the primary task with probability $w$, and for the auxiliary task with $1-w$. The primary task is trained until convergence and weight $w$ 
determines the frequency of parameter updates for the auxiliary task.
\section{Imagination Model}

\subsection{Shared Encoder}\label{sec:model:encoder}

The encoder network of our model learns a representation of a
sequence of $N$ tokens $x_{1 \ldots n}$ in the source language with a
bidirectional recurrent neural network \cite{Schuster1997}. This representation is shared between the different tasks. 
Each token is represented by a one-hot vector $\mathbf{x_i}$, which is mapped into an embedding $\mathbf{e_i}$ through a learned matrix
$\mathbf{E}$:
\begin{align}
	\mathbf{e_i} = \mathbf{x_i} \cdot \mathbf{E}
\end{align}
A sentence is processed by a pair of recurrent neural networks, where one captures the sequence left-to-right (forward), and the other captures the sequence right-to-left (backward). The initial state of the encoder $\mathbf{h_{-1}}$ is a learned parameter:
\begin{align}
	\overrightarrow{\mathbf{h_i}} = \overrightarrow{\text{RNN}}(\overrightarrow{\mathbf{h_{i-1}}}, \mathbf{e_i}) \\
    \overleftarrow{\mathbf{h_i}} = \overleftarrow{\text{RNN}}(\overleftarrow{\mathbf{h_{i-1}}}, \mathbf{e_i})
\end{align}
Each token in the source language input sequence is represented by a concatenation of the forward and backward hidden state vectors:
\begin{align}
	\mathbf{h_i} = [\overrightarrow{\mathbf{h_i}}; \overleftarrow{\mathbf{h_i}}]
\end{align}
\subsection{Neural Machine Translation Decoder}

The translation model decoder is an attention-based recurrent neural network \cite{Bahdanau2015}. Tokens in the decoder are 
represented by a one-hot vector $\mathbf{y_j}$, which is mapped into an embedding  $\mathbf{e_j}$ through a learned matrix $\mathbf{E_y}$:
\begin{align}
	\mathbf{e_j} = \mathbf{y_j} \cdot \mathbf{E_y} 
\end{align}
The inputs to the decoder are the previously predicted token $\mathbf{y_{j-1}}$, the previous decoder state $\mathbf{d_{j-1}}$, 
and a timestep-dependent context vector $\mathbf{c_j}$ calculated over the encoder hidden states:
\begin{align}
    \mathbf{d_j} = \text{RNN}(\mathbf{d_{j-1}}, \mathbf{y_{j-1}},
    \mathbf{e_j})
\end{align}
The initial state of the decoder $\mathbf{d_{\text{-1}}}$ is a nonlinear transform of the mean of the encoder states, where $\mathbf{W}_{init}$ is a learned parameter:
\begin{align}
	\mathbf{d_{\text{-1}}} = \text{tanh}(\mathbf{W}_{init} \cdot \frac{1}{N} \sum_{i}^{N}
    \mathbf{h_i})\label{eqn:decoder_init}
\end{align}
The context vector $c_j$ is a weighted sum over the encoder hidden states, where $N$ denotes the length of the source sentence:
\begin{align}
	\mathbf{c_j} = \sum_{i=1}^{N} \alpha_{ji}\mathbf{h_i} 
\end{align}
The $\alpha_{ji}$ values are the proportion of which the encoder hidden state vectors $\mathbf{ h_{1 \ldots n}}$ contribute to the decoder hidden state 
when producing the $j$th token in the translation. They are computed by a feed-forward neural network, where $\mathbf{v_a}$, 
$\mathbf{W_a}$ and $\mathbf{U_a}$ are learned parameters:
\begin{align}
	\alpha_{ji} &= \frac{\text{exp}(e_{ji})}{\sum_{l=1}^{N} \text{exp}(e_{li})}\\[1ex]
	e_{ji} &= \mathbf{v_a} \cdot \text{tanh}( \mathbf{W_{a}} \cdot \mathbf{d_{j-1}}  + \mathbf{U_{a}}  \cdot \mathbf{h_i} )
\end{align}
From the hidden state $\mathbf{d_{j}}$ the network predicts the conditional distribution of the next token $y_{j}$, given a target 
language embedding $\mathbf{e_{j-1}}$ of the previous token, the current hidden state $\mathbf{d_j}$, and the calculated context vector 
$\mathbf{c_j}$ . Note that at training time, $y_{j-1}$ is the 
true observed token; whereas for unseen data we use the inferred token $\hat{y}_{j-1}$ sampled from the output of the softmax:
\begin{align}
p(y_{j}|y_{<j}, c) = \text{softmax}(\text{tanh}( \mathbf{e_{j-1}}  + \mathbf{d_j}  + \mathbf{c_j} ))
\end{align}

The translation model is trained to minimise the negative log likelihood of predicting the target language output:
\begin{align}
  J_{NLL}(\theta, \phi^t) = - \sum_{j} \text{log p}(y_j|y_{<j}, x)
\end{align}

\subsection{Imaginet Decoder}

The image prediction decoder is trained to predict the visual feature vector of the image associated with a sentence \cite{Chrupala2015}. 
It encourages the shared encoder to learn grounded representations for the source language. 

We represent a sentence by first producing a sequence of the concatenated hidden state vectors in the Encoder (Section \ref{sec:model:encoder}). Then, the same way we initialise the hidden state of the decoder (Eqn. \ref{eqn:decoder_init}), we represent a source language sentence as the mean of the encoder states. This is the input to a feed-forward neural network that predicts the visual feature vector $\mathbf{\hat{v}}$ associated with a sentence with parameters $\mathbf{W_{vis}}$:
\begin{align}
	\mathbf{\hat{v}}  = \text{tanh}( \mathbf{W_{vis}} \cdot \frac{1}{N} \sum_{i}^{N} 
\mathbf{h_i})
\end{align}


This decoder is trained to predict the true image vector $\mathbf{v}$ with a margin-based objective, parameterised by the minimum margin
$\alpha$, and the cosine distance $d(\cdot, \cdot)$. The contrastive examples $\mathbf{v'}$ are drawn from the other instances in a minibatch:
\begin{align}
	\begin{split}
	J_{MAR}(\theta, \phi^t) = \sum_{ \mathbf{v'} \neq \mathbf{v}}
    \text{max}\{0, \alpha & - d( \mathbf{\hat{v}}, \mathbf{v}) \\[-2ex]
    & + d( \mathbf{\hat{v}} , \mathbf{v'})\}\label{eqn:imaginet}
    \end{split}
\end{align}
\section{Data}\label{sec:data}

\begin{table}
\renewcommand{\arraystretch}{1.3}
\centering
\begin{tabular}{ccccc}
\toprule
& Size & Tokens & Types & Images\\
\midrule
\multicolumn{5}{l}{Multi30K: parallel text with images}\\
En & \multirow{2}{*}{31K} & 377K & 10K & \multirow{2}{*}{31K}\\
De & & 368K & 16K & \\
\midrule
\multicolumn{5}{l}{MS COCO: external described images}\\
En    & 414K & 4.3M & 24K & 83K \\
\midrule
\multicolumn{5}{l}{News Commentary: external parallel text}\\
En   & \multirow{2}{*}{240K} & 8.31M & \multirow{2}{*}{17K} &  --\\
De   & & 8.95M & & --\\
\bottomrule
\end{tabular}
\caption{The datasets used in our experiments.}\label{tab:datasets}
\end{table}

We evaluate our model using the benchmark Multi30K dataset \cite{ElliottFrankSimaanSpecia2016}, which is the largest collection of images paired with sentences in multiple languages. This dataset contains 31,014 images paired with an English language sentence and a German language translation: 29,000 instances are reserved for training, 1,014 for development, and 1,000 for evaluation.\footnote{Multi30K also contains 155K independently collected descriptions for German and English. We do not use this data.}

The English and German sentences are preprocessed by normalising the punctuation, lowercasing and tokenizing the text using the Moses toolkit. We additionally decompound the German text  using Zmorge \cite{Sennrich2014}. This results in vocabulary sizes of 10,214 types for English and 16,022 for German.

We also use two external datasets to evaluate our model: the MS COCO dataset of English described images \cite{Chen2015}, and the English-German News Commentary parallel corpus \cite{Tiedemann2012}. When we perform experiments with the News Commentary corpus, we first calculate a 17,597 sub-word vocabulary using SentencePiece \cite{Schuster2012} over the concatentation of the Multi30K and News Commentary datasets. This gives us a shared vocabulary for the external data that reduces the number of out-of-vocabulary tokens. 

Images are represented by 2048D vectors extracted from the `pool5/7x7\_s1' layer of the GoogLeNet v3 CNN \cite{Szegedy2015}.
\section{Experiments}

We evaluate our multitasking approach with in- and out-of-domain resources. We start by reporting results of models trained using only the Multi30K dataset. We also report the results of training the {\sc imaginet} decoder with the COCO dataset. Finally, we report results on incorporating the external News Commentary parallel text into our model. Throughout, we report performance of the  En$\rightarrow$De translation using Meteor \cite{denkowski:lavie:meteor-wmt:2014} and BLEU \cite{Papineni:2002:BMA:1073083.1073135} against lowercased tokenized references.

\subsection{Hyperparameters}

The encoder is a 1000D Gated Recurrent Unit bidirectional recurrent neural network \cite[GRU]{Cho2014} with 620D embeddings. We share all of the encoder parameters between the primary and auxiliary task. The translation decoder is a 1000D GRU recurrent neural network, with a 2000D context vector over the encoder states, and 620D word embeddings \cite{Sennrich2017}. The Imaginet decoder is a single-layer feed-forward network, where we learn the parameters $\mathbf{W_{vis}} \in \mathbb{R}^{2048\text{x}2000}$ to predict the true image vector with $\alpha = 0.1$ for the Imaginet objective (Equation \ref{eqn:imaginet}). The models are trained using the Adam optimiser with the default hyperparameters \cite{Kingma2015} in minibatches of 80 instances. The translation task is defined as the primary task and convergence is reached when BLEU has not increased for five epochs on the validation data. Gradients are clipped when their norm exceeds 1.0. Dropout is set to 0.2 for the embeddings and the recurrent connections in both tasks \cite{Gal2016}. Translations are decoded using beam search with 12 hypotheses.

\subsection{In-domain experiments}\label{sec:in_domain_all}


\begin{table}
\centering
\renewcommand{\arraystretch}{1.3}
\begin{tabular}{lccc}
\toprule
& Meteor & BLEU \\
\midrule
NMT & 54.0 $\pm$ 0.6 & 35.5 $\pm$ 0.8\\
\newcite{Calixto2017c} & 55.0 & 36.5 \\
\newcite{Calixto2017b} & 55.1 & 37.3 \\
Imagination & 55.8 $\pm$ 0.4 & 36.8 $\pm$ 0.8\\
\newcite{toyama2016neural} & 56.0  & 36.5 \\
\newcite{Hitschler2016} & 56.1 & 34.3 \\
Moses & 56.9 & 36.9\\
\bottomrule
\end{tabular}
\caption{En$\rightarrow$De translation results on the Multi30K dataset. Our Imagination model is competitive with the state of the art when it is trained on in-domain data. We report the mean and standard deviation of three random initialisations.}
\label{tab:results:in-domain}
\end{table}

We start by presenting the results of our multitask
model trained using only the Multi30K dataset. We compare against state-of-the-art approaches and text-only baselines. Moses is the phrase-based machine translation model \cite{Koehn2007} reported in \cite{Specia2016}. NMT is a text-only neural machine translation model.
\newcite{Calixto2017c} is a double-attention model over the source language and the image. 
\newcite{Calixto2017b} is a multimodal translation model that conditions the decoder on semantic image vector extracted from the VGG-19 CNN. 
\newcite{Hitschler2016} uses visual features in a target-side retrieval model for translation. \newcite{toyama2016neural} is most comparable to our approach: it is a multimodal variational NMT model that infers latent variables to represent the source language semantics from the image and linguistic data.
 
Table \ref{tab:results:in-domain} shows the results of this experiment.
We can see that the combination of the attention-based translation model and the image
prediction model is a 1.8 Meteor point improvement over the NMT
baseline, but it is 1.1 Meteor points worse than the strong Moses baseline. Our approach is competitive with previous approaches that use visual features as inputs to the decoder and the target-side reranking model. It also competitive with \newcite{toyama2016neural}, which also only uses images for training.  These results confirm that our multitasking approach uses the image prediction task to improve the encoder of the translation model.

\begin{table}[t]
\centering
\renewcommand{\arraystretch}{1.3}
\begin{tabular}{lcc}
\toprule
& Meteor & BLEU \\
\midrule
Imagination & 55.8 $\pm$ 0.4 & 36.8 $\pm$ 0.8 \\
Imagination (COCO) & 55.6 $\pm$ 0.5 & 36.4 $\pm$ 1.2 \\
\bottomrule
\end{tabular}
\caption{Translation results when using out-of-domain described images. Our approach is still effective when the image prediction model is trained over the COCO dataset.}\label{tab:results:ood-coco}
\end{table}

\subsection{External described image data}\label{sec:experiments:ood-images}

\begin{table}
\centering
\renewcommand{\arraystretch}{1.3}
\begin{tabular}{@{}lcc}
\cmidrule[.08em](r{1ex}){1-3}
 & Meteor & BLEU \\
\cmidrule[.05em](r{1ex}){1-3}
NMT & 52.8 $\pm$ 0.6 & 33.4 $\pm$ 0.6\\
+ NC & 56.7 $\pm$ 0.3 & 37.2 $\pm$ 0.7 \\
+ Imagination & 56.7 $\pm$ 0.1 & 37.4 $\pm$ 0.3 \\
+ Imagination (COCO) & 57.1 $\pm$ 0.2 & 37.8 $\pm$ 0.7 \\
\newcite{Calixto2017c} & 56.8 & 39.0 \\
\cmidrule[.08em](r{1ex}){1-3}
\end{tabular}
	\caption{Translation results with out-of-domain parallel text and described images. We find further improvements when we multitask with the News Commentary (NC) and COCO datasets.}\label{tab:results:ood-both}
\end{table}

Recall from Section \ref{sec:problem} that we are interested in scenarios where $x$, $y$, and $v$ are drawn from different sources. We now experiment with separating the translation data from the described image data using $\mathcal{D}_{image}$: MS COCO dataset of 83K described images\footnote{Due to differences in the vocabularies of the repsective datasets, we do not train on examples where more than 10\% of the tokens are out-of-vocabulary in the Multi30K dataset.} and $\mathcal{D}_{text}$: Multi30K parallel text.

Table \ref{tab:results:ood-coco} shows the results of this experiment. 
We find that there is no significant difference between training the {\sc imaginet} decoder on in-domain (Multi30K) or out-of-domain data (COCO). This result confirms that we can separate the parallel text from the described images.

\subsection{External parallel text data}

\begin{table*}[t]
\centering
	\begin{tabular}{ccccccc}
    \toprule
    & \multicolumn{2}{c}{Parallel text} & \multicolumn{2}{c}{Described images} \\
    & Multi30K & News Commentary & Multi30K & COCO & Meteor & BLEU \\
    \midrule
    \parbox[t]{2mm}{\multirow{3}{*}{\rotatebox[origin=c]{90}{Zmorge}}} & \checkmark   &   &   &       & 56.2   & 37.8 \\[2.5ex]
    & \checkmark   &   &  \checkmark  &           & 57.6   & 39.0 \\
    \midrule
    \parbox[t]{2mm}{\multirow{4}{*}{\rotatebox[origin=c]{90}{Sub-word}}} & \checkmark   &   &   &       & 54.4   & 35.0 \\
	& \checkmark   &   \checkmark    &   &               & 58.6   & 39.4 \\
    & \checkmark   & \checkmark  & \checkmark   &   & 59.0   & 39.5 \\
    & \checkmark   &  \checkmark &   &   \checkmark      & \textbf{59.3}   & \textbf{40.2} \\
\bottomrule
\end{tabular}
    \caption{Ensemble decoding results. Zmorge denotes models trained with decompounded German words; Sub-word denotes joint SentencePiece word splitting (see Section \ref{sec:data} for more details).}\label{tab:results:ensemble}
\end{table*}

We now experiment with training our model on a combination of the Multi30K and the News Commentary English-German data. In these experiments, we concatenate the Multi30K and News Commentary datasets into a single $\mathcal{D}_{text}$ training dataset, similar to \newcite{Freitag2016}. We compare our model against \newcite{Calixto2017c}, who pre-train their model on the WMT'15 English-German parallel text and back-translate \cite{Sennrich2016b} additional sentences from the bilingual independent descriptions in the Multi30K dataset (Footnote 2).


Table \ref{tab:results:ood-both} presents the results. The text-only NMT model using sub-words is 1.2 Meteor points lower than decompounding the German text. Nevertheless, the model trained over a concatentation of the parallel texts is a 2.7 Meteor point improvement over this baseline (+ NC) and matches the performance of our Multitasking model that uses only in-domain data (Section \ref{sec:in_domain_all}). We do not see an additive improvement for the multitasking model with the concatenated parallel text and the in-domain data (+ Imagination) using a training objective interpolation of $w = 0.89$ (the ratio of the training dataset sizes). This may be because we are essentially learning a translation model and the updates from the {\sc imaginet} decoder are forgotten. Therefore, we experiment with multitasking the concatenated parallel text and the COCO dataset ($w=0.5$). We find that balancing the datasets improves over the concatenated text model by 0.4 Meteor (+ Imagination (COCO)). Our multitasking approach improves upon Calixto et al. by 0.3 Meteor points. Our model can be trained in 48 hours using 240K parallel sentences and 414K described images from out-of-domain datasets. Furthermore, recall that our model does not use images as an input for translating unseen data, which results in 6.2\% fewer parameters compared to using the 2048D Inception-V3 visual features to initialise the hidden state of the decoder.

\begin{table*}[h]
\renewcommand{\arraystretch}{1.2}
\begin{subtable}{0.22\textwidth}
\centering
\begin{tabular}{c}
\includegraphics[width=0.7\textwidth]{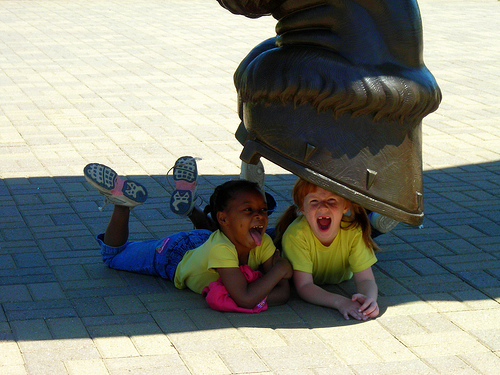}
\end{tabular}
\end{subtable}%
\begin{subtable}{0.75\textwidth}
\begin{tabular}{rp{29em}}
Source: & two children on their stomachs lay on the ground under a pipe \\
NMT: & zwei kinder \textcolor{red}{auf ihren gesichtern} liegen unter dem boden auf dem boden \\
Ours: & zwei kinder liegen bäuchlings auf dem boden unter einer \textcolor{red}{schaukel}
 \\
\end{tabular}
\end{subtable}

\vspace{0em}
 
\begin{subtable}{0.22\textwidth}
\centering
\begin{tabular}{c}
\includegraphics[width=0.7\textwidth]{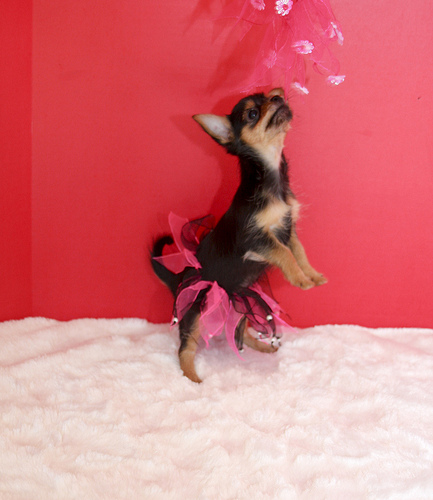}
\end{tabular}
\end{subtable}%
\begin{subtable}{0.75\textwidth}
\begin{tabular}{rp{29em}}
Source: & small dog in costume stands on hind legs to reach dangling flowers \\
NMT: & ein kleiner hund steht auf dem hinterbeinen und \textcolor{red}{läuft} , \textcolor{red}{nach links von blumen zu sehen}  \\
Ours: & ein kleiner hund in einem kostüm steht auf den hinterbeinen , um die blumen zu erreichen\\
\end{tabular}

\end{subtable}

\vspace{0em}

\begin{subtable}{0.22\textwidth}
\centering
\begin{tabular}{c}
\includegraphics[width=0.7\textwidth]{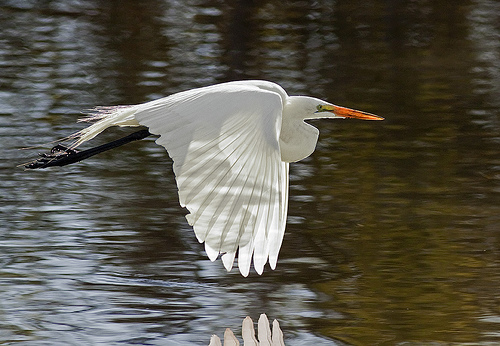}
\end{tabular}
\end{subtable}%
\begin{subtable}{0.75\textwidth}
\begin{tabular}{rp{29em}}
Source: & a bird flies across the water \\
NMT: & ein vogel fliegt über das wasser \\
Ours: & ein vogel fliegt \textcolor{red}{durch} das wasser\\
\end{tabular}
\end{subtable}

\caption{Examples where our model improves or worsens the translation compared to the NMT baseline. Top: NMT translates the wrong body part; both models skip ``pipe''. Middle: NMT incorrectly translates the verb and misses several nouns. Bottom: Our model incorrectly translates the preposition.}\label{tab:results:examples}
\end{table*}

\subsection{Ensemble results}

Table \ref{tab:results:ensemble} presents the results of ensembling different randomly initialised models. We achieve a start-of-the-art result of 57.6 Meteor for a model trained on only in-domain data. The improvements are more pronounced for the models trained using sub-words and out-of-domain data. An ensemble of baselines trained on sub-words is initially worse than an ensemble trained on Zmorge decompounded words. However, we always see an improvement from ensembling models trained on in- and out-of-domain data. Our best ensemble is trained on Multi30K parallel text, the News Commentary parallel text, and the COCO descriptions to set a new state-of-the-art result of 59.3 Meteor. 

\subsection{Qualitative Examples}

Table \ref{tab:results:examples} shows examples of where the multitasking model improves or worsens translation performance compared to the baseline model\footnote{We used MT-ComparEval \cite{Klejch2015}}. The first example shows that the baseline model makes a significant error in translating the pose of the children, translating ``on their stomachs'' as ``on their faces''). The middle example demonstrates that the baseline model translates the dog as walking (``läuft'') and then makes grammatical and sense errors after the clause marker. Both models neglect to translate the word ``dangling'', which is a low-frequency word in the training data. There are instances where the baseline produces better translations than the multitask model: In the bottom example, our model translates a bird flying through the water (``durch'') instead of ``over'' the water.
\section{Discussion}

\subsection{Does the model learn grounded representations?}

\begin{figure*}
 \begin{subfigure}[b]{1\textwidth}
 \centering
\includegraphics[width=0.75\textwidth]{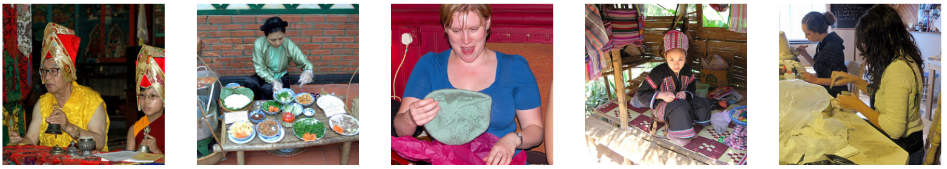}
\caption{Nearest neighbours for ``a native woman is working on a craft project .''}
\vspace{2ex}
\end{subfigure}
 \begin{subfigure}[b]{1\textwidth}
 \centering
\includegraphics[width=0.75\textwidth]{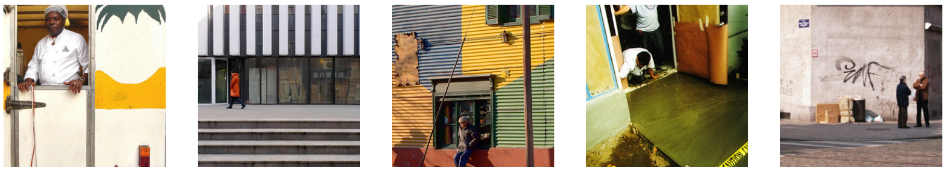}
\caption{Nearest neighbours for ``there is a cafe on the street corner with an oval painting on the side of the building .''}
\end{subfigure}
\caption{We can interpret the {\sc imaginet} Decoder by visualising the predictions made by our model.}\label{fig:discussion:examples}
\end{figure*}

A natural question to ask if whether the multitask model is actually learning representations that are relevant for the images. We answer this question by evaluating the Imaginet decoder in an image--sentence ranking task. Here the input is a source language sentence, from which we predict its image vector $\mathbf{\hat{v}}$. The predicted vector $\mathbf{\hat{v}}$ can be compared against the true image vectors $\mathbf{v}$ in the evaluation data using the cosine distance to produce a ranked order of the images. Our model returns a median rank of 11.0 for the true image compared to the predicted image vector. 
Figure \ref{fig:discussion:examples} shows examples of the nearest neighbours of the images predicted by our multitask model. We can see that the combination of the multitask source language representations and {\sc imaginet} decoder
leads to the prediction of relevant images. This confirms that the shared encoder is indeed learning visually grounded representations.

\subsection{The effect of visual feature vectors}

We now study the effect of varying the Convolutional Neural Network used to extract the visual features used in the Imaginet decoder. 
It has previously been shown that the choice of visual features can affect the performance of vision and language models \cite{Jabri2016,Kiela2016}. 
We compare the effect of training the {\sc imaginet} decoder to predict different types of image features, namely: 4096D features extracted from the `fc7`' layer of the VGG-19 model \cite{Simonyan2015}, 2048D features extracted from the `pool5/7x7\_s1' layer of InceptionNet V3 \cite{Szegedy2015}, and 2048D features extracted from `avg\_pool` layer of ResNet-50 \cite{He2016}. 
Table \ref{tab:results:features} shows the results of this experiment. 
There is a clear difference between predicting the 2048D vectors (Inception-V3 and ResNet-50) compared to the 4096D vector from VGG-19). 
This difference is reflected in both the translation Meteor score and the Median rank of the images in the validation dataset. This is likely because it is easier to learn the parameters of the image prediction model that has fewer parameters (8.192 million for VGG-19 vs. 4.096 million for Inception-V3 and ResNet-50).
However, it is not clear why there is such a pronounced difference between the Inception-V3 and ResNet-50 models\footnote{We used pre-trained CNNs (\url{https://github.com/fchollet/deep-learning-models}), which claim equal ILSVRC object recognition performance for both models: 7.8\% top-5 error with a single-model and single-crop.}. 
Further work is needed to understand the difference in these results.

\begin{table}
\centering
\renewcommand{\arraystretch}{1.3}
\begin{tabular}{lcc}
\toprule
& Meteor & Median Rank \\
\midrule
Inception-V3 & 56.0 $\pm$ 0.1 & 11.0 $\pm$ 0.0\\
Resnet-50 & 54.7 $\pm$ 0.4 & 11.7 $\pm$ 0.5 \\
VGG-19 & 53.6 $\pm$ 1.8 & 13.0 $\pm$ 0.0 \\
\bottomrule
\end{tabular}
\caption{The type of visual features predicted by the {\sc imaginet} Decoder has a strong impact on the Multitask model performance.}\label{tab:results:features}
\end{table}
\section{Related work}



Initial work on multimodal translation has focused on approaches that
use either semantic or spatially-preserving image features as inputs to a translation model. Semantic image features are typically extracted from the final pooling
layer of a pre-trained object recognition CNN, e.g.
`pool5/7x7\_s1' in GoogLeNet \cite{Szegedy2015}. This type of feature vector has been used
conditioning input to the encoder \cite{ElliottFrankHasler2015,Huang2016}, in
the decoder \cite{Libovicky2016}, or as additional features in a phrase-based
translation model \cite{Shah2016,Hitschler2016}.
Spatially-preserving image features are extracted from deeper inside a CNN, where the position of a feature in the tensor is related to its position in the image. These features have
been used in ``double-attention models'', which calculate independent context vectors of the source language hidden states and a
convolutional image feature map
\cite{Calixto2016,Caglayan2016b,Calixto2017c}.
Similar to most of these approaches, we use an attention-based translation model, but our multitask model does not use images for translation.

More related to our work are the recent papers by \newcite{toyama2016neural}, \newcite{Saha2016}, and \newcite{Nakayama2016}. \newcite{toyama2016neural} extend the Variational Neural Machine Translation model \cite{zhang2016variational} by inferring latent variables to \emph{explicitly} model the semantics of source sentences from both image and linguistic information. Their model does not condition  on images for translation. They report improvements on the Multi30K data set when using multimodal information, however, their model adds additional parameters in the form of ``neural inferrer'' modules. In our multitask model, the grounded semantics are represented \emph{implicitly} in the hidden states of the shared encoder. Furthermore, they assume Source-Target-Image aligned training data; whereas our approach achieves equally good results if we train on separate Source-Image and Source-Target datasets.

\newcite{Saha2016} study cross-lingual image description where the task is to generate a sentence in language $L_{1}$ given the image, given only Image-$L_{2}$ and $L_{1}$-$L_{2}$ parallel corpora. 
They propose a Correlational Encoder-Decoder to model the Image-$L_{2}$ and $L_{1}$-$L_{2}$ data. Their model learns 
correlated representations for paired Image-$L_{2}$ data and decodes 
$L_{1}$ from this joint representation. Similarly to work, 
the encoder is trained by minimizing two loss functions: the Image-$L_{2}$ correlation loss, and the $L_{1}$ decoding cross-entropy loss. 

\newcite{Nakayama2016} consider a zero-resource problem
where the task is to translate from $L_{1}$ to $L_{2}$ but only 
Image-$L_{1}$ and Image-$L_{2}$ corpora are available. 
Their model embeds the image, $L_{1}$, and $L_{2}$ in a joint multimodal space learned through minimizing a multi-task ranking loss 
between both pairs of examples. The main difference between this approach and our model is that we focus on
\emph{enriching} source language representations with visual 
information, rather than addressing the zero-resource issue.

Multitask Learning improves the 
generalisability of a model by requiring it to be useful for more than one task \cite{Caruana1997}.
This approach has recently been used to improve 
the performance of sentence compression using eye gaze as an 
auxiliary task \cite{Klerke2016}, and to improve shallow parsing 
accuracy through the auxiliary task of predicting keystrokes in an 
out-of-domain corpus \cite{Plank2016}. These works hypothesise a relationship between
specific biometric measurements and specific NLP tasks motivated by cognitive-linguistic theories. 
More recently, \newcite{Bingel2017} analysed the beneficial relationships between primary and auxiliary sequential prediction tasks.
In the translation literature, multitask learning has been used to learn a one-to-many 
languages translation model \cite{Dong2015}, a multi-lingual translation model with a single attention mechanism shared across multiple languages \cite{Firat2016}, and in multitask sequence-to-sequence learning without an attention-based decoder \cite{Luong2016}. In our multitask framework we explore the benefits of grounded learning in the specific case of multimodal translation. Our model combines sequence prediction with continuous (image) vector prediction, compared to previous work on multitask learning for sequence prediction tasks.

Visual representation prediction has been tackled as an unsupervised and supervised problem. In an unsupervised setting, \newcite{Srivastava2015} propose an LSTM Autoencoder to predict video frames as a reconstruction task or as a future prediction task. In a supervised setting, \newcite{Lin2015} use a conditional random field to imagine the composition of a clip-art scene for visual paraphrasing and fill-in-the-blank tasks. \newcite{Chrupala2015} predict the image vector associated with a sentence using an L2 loss; they found this approach improves multi-modal word similarity compared to text-only baselines. \newcite{Gelderloos2016} predict the visual feature vector associated with a sequence of phonemes using a max-margin loss, similar to our image prediction objective. \newcite{Collell2017} learn to predict the visual feature vector associated with a word for word similarity and relatedness tasks; their best approach combines a word representation with the predicted image representation, which allows zero-shot learning for unseen words. \newcite{Pasunuru2017} propose a multi-task learning model for video description that combines unsupervised video frame reconstruction, lexical entailment, and video description in a single framework. They find improvements from using out-of-domain resources for the entailment and video frame prediction tasks, similar to the improvements we find from using out-of-domain parallel text and described images.
\section{Conclusion}

We decompose multimodal translation into two sub-problems: learning to translate and learning visually grounded representations. In a multitask learning framework, we show how these sub-problems can be addressed by sharing an encoder between a translation model and an image prediction model. Our approach achieves state-of-the-art results on the Multi30K dataset without using images for translation. We show that training on separate parallel text and described image datasets does not hurt performance, encouraging future research on multitasking with diverse sources of data. 
Furthermore, we still find improvements from image prediction when we improve our text-only baseline with the out-of-domain parallel text. Future work includes adapting our decomposition to other NLP tasks that may benefit from out-of-domain resources, such as semantic role labelling, dependency parsing, and question-answering; exploring methods for inputting the (predicted) image into the translation model; experimenting with different image prediction architectures; multitasking different translation languages into a single shared encoder; and multitasking in both the encoder and decoder(s).
\section*{Acknowledgments}
We thank Joost Bastings for sharing his multitasking Nematus model, Wilker Aziz for discussions about formulating the problem, Stella Frank for finding and explaining the qualitative examples to us, and Afra Alishahi, Grzegorz Chrupała,
and Philip Schulz for feedback on earlier drafts of the paper. DE acknowledges the support of NWO
Vici grant nr. 277-89-002 awarded to K. Sima’an, an Amazon Academic Research Award, and a hardware grant from the NVIDIA Corporation.

\bibliography{emnlp2017}
\bibliographystyle{eacl2017}

\end{document}